%% file: main.tex
\documentclass[letterpaper, 10 pt, conference]{URL-ieeeconf}
\IEEEoverridecommandlockouts    
\usepackage{url}
\usepackage{array,tabularx}
\usepackage{amssymb}
\usepackage{cite}
\usepackage{blindtext}
\usepackage{booktabs}
\usepackage{pifont}
\usepackage{bbding}
\usepackage{algorithmic}
\usepackage{graphicx}
\usepackage{textcomp}
\usepackage{mathtools}
\usepackage{subcaption}
\usepackage{algorithm}
\usepackage{xcolor}
\usepackage{color, soul}
\usepackage{multirow}
\usepackage{xstring}
\usepackage{hhline}
\usepackage{kotex}
\usepackage{siunitx}
\usepackage{physics}
\usepackage{tikz}
\usepackage{rotating}
\usepackage{hyperref}

\def\figref#1{Fig.~\ref{#1}}
\def\tabref#1{Table~\ref{#1}}
\def\eqref#1{(\ref{#1})}

\newcommand\vsfig{\vspace{-0.2cm}}

\newcommand{\rom}[1]{\uppercase\expandafter{\romannumeral #1\relax}}

\newcommand\eg{{e.g.}}
\newcommand\ie{{i.e.}}
\newcommand\etal{\emph{et al.}}
\def\etalcite#1{\etal~\cite{#1}}

\definecolor{rvc}{RGB}{0, 0, 0}

\title{\LARGE \bf Similar but Different: A Survey of Ground Segmentation and Traversability Estimation for Terrestrial Robots}

\author{Hyungtae Lim$^{1}$, \textit{Member, IEEE}, Minho Oh$^{1}$, \textit{Student Member, IEEE}, Seungjae Lee$^{1}$, \textit{Student Member, IEEE},\\ Seunguk Ahn$^2$, and Hyun Myung$^{1*}$, \textit{Senior Member, IEEE} %
  \thanks{$^*$Corresponding author: Hyun Myung}
  \thanks{$^{1}$Hyungtae Lim, Minho Oh, Seungjae Lee, and Hyun Myung are with the School of Electrical Engineering, KAIST (Korea Advanced Institute of Science and Technology), Daejeon, 34141, Republic of Korea. {\tt\scriptsize \{shapelim, minho.oh, sj98lee, hmyung\}@kaist.ac.kr}}
  \thanks{$^2$Seunguk Ahn is with the Hanwha Aerospace, 6, Pangyo-ro 319~beon-gil, Bundang-gu, Seongnam-si, Gyeonggi-do, Republic of Korea. {\tt\scriptsize seunguk.ahn@hanwha.com} \hfill \break
  \indent This project was supported by the grant from Hanwha Aerospace as part of the development of autonomous driving technology for unstructured environments. The students are supported by BK21 FOUR~(Republic of Korea).}
}

\begin{document}
\maketitle
\thispagestyle{empty}
\pagestyle{empty}

\begin{abstract}
With the increasing demand for mobile robots and autonomous vehicles, several approaches for long-term robot navigation have been proposed.
Among these techniques, ground segmentation and traversability estimation play important roles in perception and path planning, respectively.
Even though these two techniques appear similar, their objectives are different.
Ground segmentation divides data into ground and non-ground elements;
thus, it is used as a preprocessing stage to extract objects of interest by rejecting ground points.
In contrast, traversability estimation identifies and comprehends areas in which robots can move safely.
Nevertheless, some researchers use these terms without clear distinction, leading to misunderstanding the two concepts.
Therefore, in this study, we survey related literature and clearly distinguish ground and traversable regions considering four aspects:
a)~maneuverability of robot platforms, b)~position of a robot in the surroundings, c)~subset relation of negative obstacles, and d)~subset relation of deformable objects.

\end{abstract}

\section{INTRODUCTION}

With the increasing demand for mobile robots and autonomous vehicles, several approaches for long-term robot navigation have been proposed~\cite{lim2021erasor, chen2021ral, sung2021if, lim2023erasor2, pomerleau2014long}.
Specifically, to enable such terrestrial robots to autonomously perceive surroundings, localize themselves on the map, and then navigate toward their destination without any human supervision,
lots of techniques have been extensively studied, such as object recognition~\cite{yoon2019crv, oh2022travel, chen2022ral}, visual place recognition~\cite{zeng2017enhancing, chang2020spoxelnet, lee20232, arshad2023robust, park2020robust, lee2019bag}, simultaneous localization and mapping~(SLAM)~\cite{choi2011localization, park2011localization, shan2018lego, lim2020normal, shan2021icra, seo2022pago, song2022dynavins, song2022g2p, lim2023adalio, hata2015feature, stachniss2005mobile, sarlin2019cvpr}, registration~\cite{chen1992ivc, lim2022asinlge, eckart2018hgmr, lim2023quatropp}, and path planning/following~\cite{gennery1999traversability, wermelinger2016navigation, lee2021real, lee2022faro, yu2023smooth, liu2020robotic}.

\begin{figure}[t!]
	\centering
	\captionsetup{font=footnotesize}
	\begin{subfigure}[b]{0.48\textwidth}
		\includegraphics[width=1.0\textwidth]{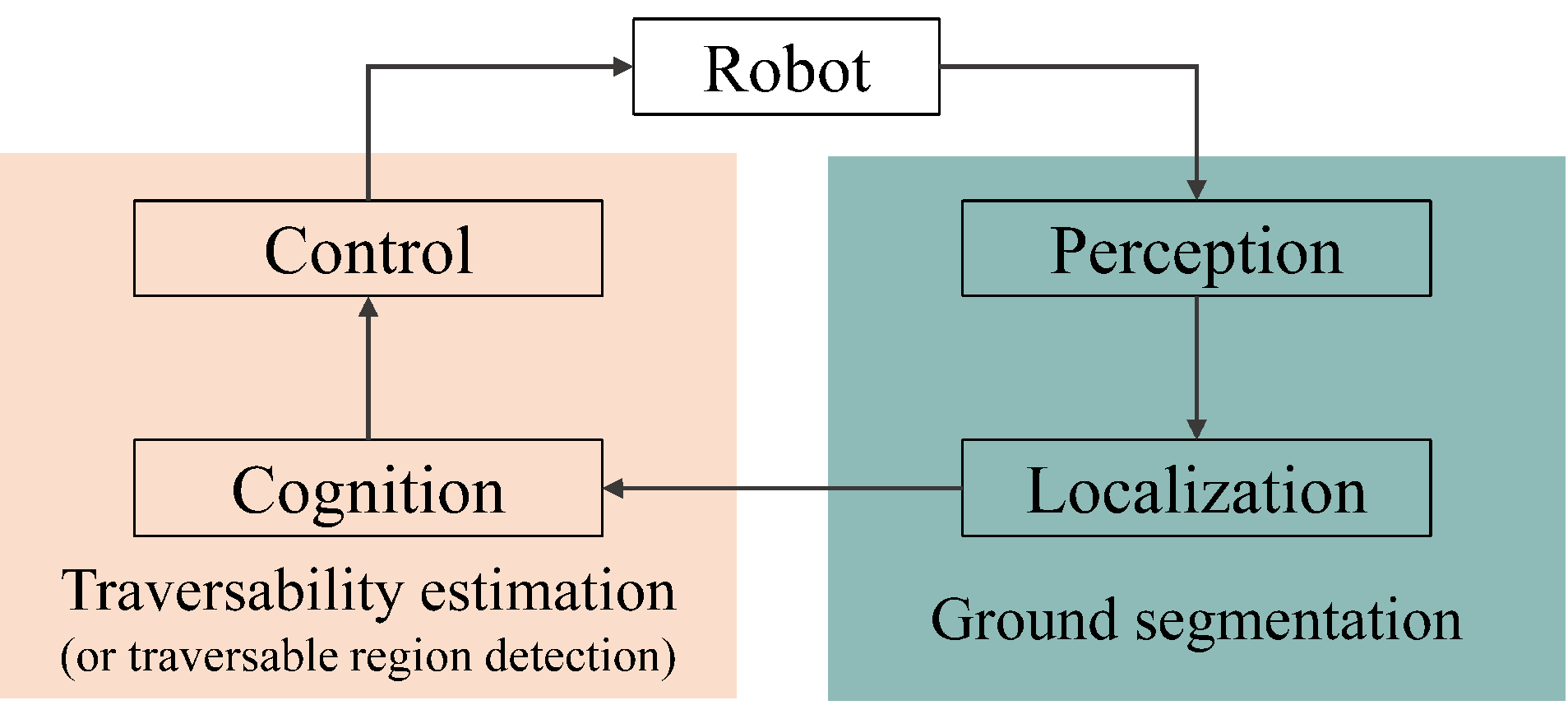}
	\end{subfigure}
	\vspace{-0.1cm}
	\caption{Diagram of robot navigation with ground segmentation and traversability estimation. Note that ground segmentation is a technique for perceiving and mapping surroundings, whereas traversability estimation is aimed at cognition and motion control~(best viewed in color).}
	\label{fig:objective}
	\vsfig
\end{figure}

Among these techniques, \textit{ground segmentation} and \textit{traversability estimation~(or traversable region detection)} play important roles in the perception and path planning, respectively~\cite{kim2023road, xue2023multiscale, cui2009floor, pae2021track, park2013terrain, park2012sampling}.
Even though these two techniques appear similar, their objectives are different.
Ground segmentation divides data into ground and non-ground elements;
thus, it is used as a preprocessing stage to extract objects of interest by rejecting ground points.
In contrast, traversability estimation identifies and comprehends areas in which robots can move safely~\cite{shan2018bayesian, suryamurthy2019terrain}.
Therefore, as shown in~\figref{fig:objective}, which illustrates the robot navigation architecture, ground segmentation is required at the perception level,
whereas traversability estimation is required at the cognition level.

\begin{figure}[t!]
	\centering
	\captionsetup{font=footnotesize}
	\begin{subfigure}[b]{0.48\textwidth}
		\includegraphics[width=1.0\textwidth]{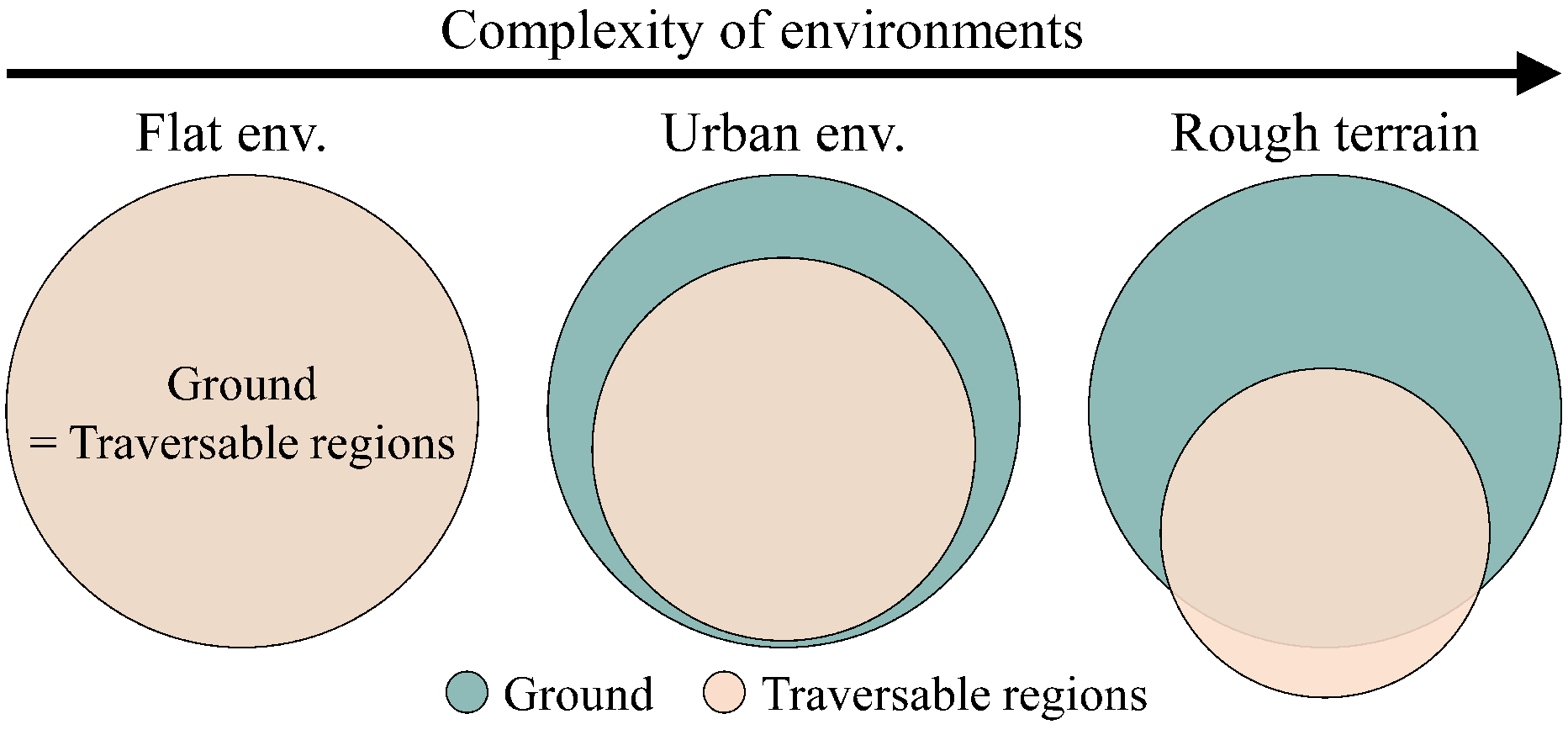}
	\end{subfigure}
	\vspace{-0.1cm}
	\caption{Subset relation between ground and traversable regions depending on the complexity of environments considering wheeled mobile robots. Note that in rough terrain environments, some parts of traversable regions may not be included in the ground owing to deformable objects~(best viewed in color).}
	\label{fig:categorization}
	\vsfig
\end{figure}

These two terms, traversable region and ground, did not need to be significantly differentiated in the 20th century because most robots were wheeled robots employed in flat environments, such as factories or laboratories.
But, recently, service robots with legged locomotion, such as bipedal or quadrupedal robots, are being developed~\cite{hwangbo2019learning, miki2022learning, nahrendra2023dreamwaq, nahrendra2023robust} which can be employed in more complex environments, ranging from urban environments to rough terrain and even disaster situations~\cite{hu2018multi}.
Thus, the categories of ground and traversable regions have become more complicated, as shown in~\figref{fig:categorization}.
Consequently, some researchers use these terms ambiguously and without clear distinction, leading to misunderstanding these concepts~\cite{byun2015drivable, na2016drivable, oh2022travel, fu2022rs, xue2023multiscale}.

Therefore, in this study, we aim to provide clear definitions of ground and traversable regions to alleviate confusion in terminology.
While several well-written surveys exist regarding ground segmentation~\cite{gomes2023sensors} and traversability estimation~\cite{papadakis2013terrain, sevastopoulos2022survey, borges2022survey, beycimen2023survey}, respectively,
these survey papers are less focused on clarifying ground and traversable regions.
Thus, we do not only list state-of-the-art algorithms but also provide comprehensive explanation of their definitions and clarifications from various perspectives.
Particularly, we discuss how the categories change depending on the following four criteria: a)~maneuverability of robot platforms, b)~the position of a robot in the surroundings, c)~presence of negative obstacles~\cite{borges2022survey}, \eg~potholes and pits, and d) deformable objects,~\eg~lawns, bushes, and reeds.

The remainder of this article is organized as follows.
In Sections~\rom{2} and \rom{3}, the definition and objective of ground segmentation and traversable regions are explained, respectively.
In Section~\rom{4}, we present explicit differences between the ground and traversable regions based on the four criteria described above.
Finally, Section~\rom{5} concludes this study.

\section{GROUND SEGMENTATION}

\newcommand{\pointcloud}{\mathcal{P}}
\subsection{Objective of Ground Segmentation}\label{subsec:obj_ground_seg}

\begin{figure}[t!]
	\centering
	\captionsetup{font=footnotesize}
	\begin{subfigure}[b]{0.40\textwidth}
		\includegraphics[width=1.0\textwidth]{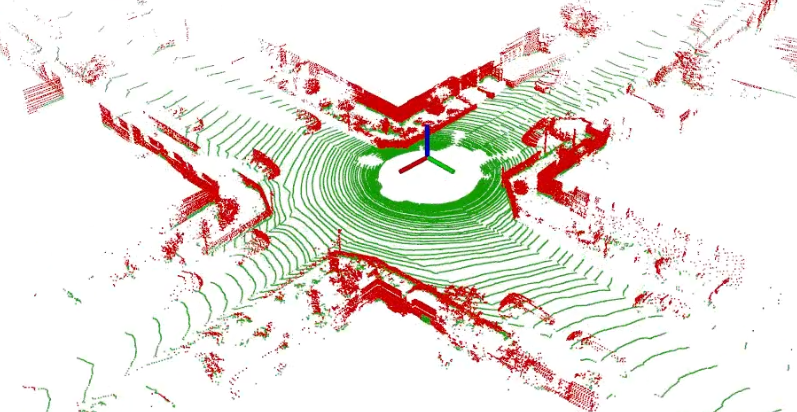}
	\end{subfigure}
	\vspace{-0.1cm}
	\caption{Example of estimated ground and non-ground points by the ground segmentation approach, Patchwork++~\cite{lee2022patchworkpp}. The green and red points denote the estimated ground and non-ground points, respectively~(best viewed in color).}
	\label{fig:patchworkpp}
	\vsfig
\end{figure}

First, we provide a clear definition of ground segmentation.
Compared with traversability estimation, the output of ground segmentation is straightforward.
Ground segmentation discriminates between the ground and non-ground regions from the input data, as shown in \figref{fig:patchworkpp}.
In this paper, we place more emphasis on the point cloud data as an input.

Formally, we define the data captured by a sensor for each frame as $\pointcloud$, which could be a 3D point cloud measured by a 3D range sensor such as a LiDAR.
Denoting the data from the actual ground and non-ground regions as $G$ and $H$, respectively,
the relationships between $G$ and $H$ are defined as follows:

\begin{equation}
G \cup {H} = \pointcloud \: \; \text{and} \: \;  G \cap {H} = \varnothing.
	\label{eq:def_ground}
\end{equation}
Note that ${H}$ includes objects of interest, such as urban structures, walls, street trees, pedestrians, and vehicles.

However, misclassification may occur because the ground in urban environments is sometimes rough and locally irregular~\cite{lim2021patchwork}.
Thus, some points from non-ground objects can be included in the estimated ground points and vice versa.
Formally, we denote the estimated ground and non-ground as $\hat{G}$ and $\hat{H}$, respectively.
Then, $\hat{G}$ and $\hat{H}$ can be expressed as follows:

\vspace{-0.2cm}
\begin{equation}
\hat{G}=\text{TP}\cup\text{FP} \: \; \text{and} \: \; \hat{H}=\text{FN} \cup \text{TN}
\end{equation}

\noindent where $\hat{G}$ and $\hat{H}$ also satisfy $\hat{G} \cup {\hat{H}} = \pointcloud$, and TP, FP, FN, and TN denote sets of \textit{true positives}, \textit{false positives}, \textit{false negatives}, and \textit{true negatives}, respectively.

In summary, ground segmentation aims to discriminate between $\hat{G}$ and $\hat{H}$ given $\pointcloud$ while minimizing FPs and FNs.

As described in Section~\rom{1}, ground segmentation is an approach regarding perception; thus, the goal is to perform ground segmentation as a preprocessing step based on the observation that most terrestrial objects are inevitably in contact with the ground.
Thus, the term \textit{ground} denotes all regions where all objects, including movable objects such as terrestrial vehicles or humans, can stand.

At the perception level, ground segmentation plays two key roles. First, by filtering out the estimated ground information, ground segmentation significantly reduces the computational burden of its following procedure, such as the above-ground object segmentation~\cite{oh2022travel} or clustering~\cite{bogoslavskyi2016depthclustering}.
For example, when testing ground segmentation in the SemanticKITTI dataset~\cite{behley2019iccv}, it can reduce 50$-$60\% of the complexity of the algorithms because 50$-$60\% of the points are identified as ground points once a 3D point cloud is acquired using a 64-channel 3D LiDAR sensor~\cite{vel64, ouster0}.

\begin{figure}[t!]
	\centering
	\captionsetup{font=footnotesize}
	\begin{subfigure}{0.40\textwidth}
		\includegraphics[width=1.0\textwidth]{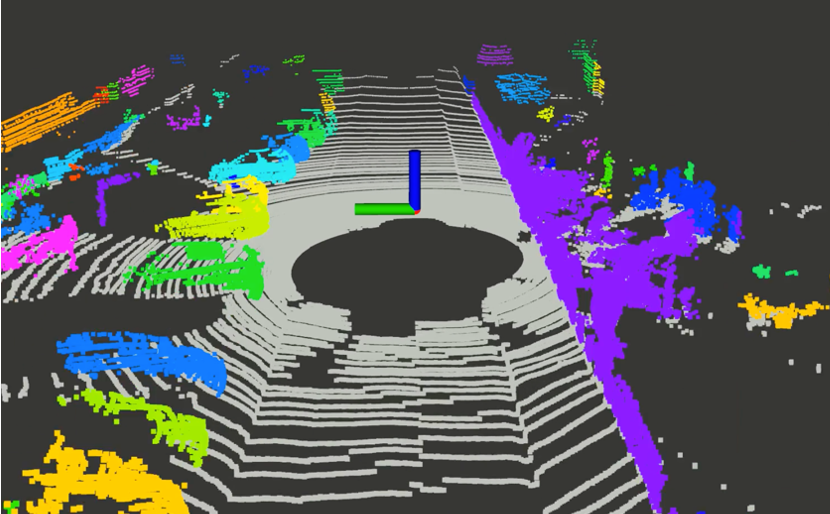}
		\caption{\centering}
	\end{subfigure}
	\begin{subfigure}{0.45\textwidth}
		\includegraphics[width=1.0\textwidth]{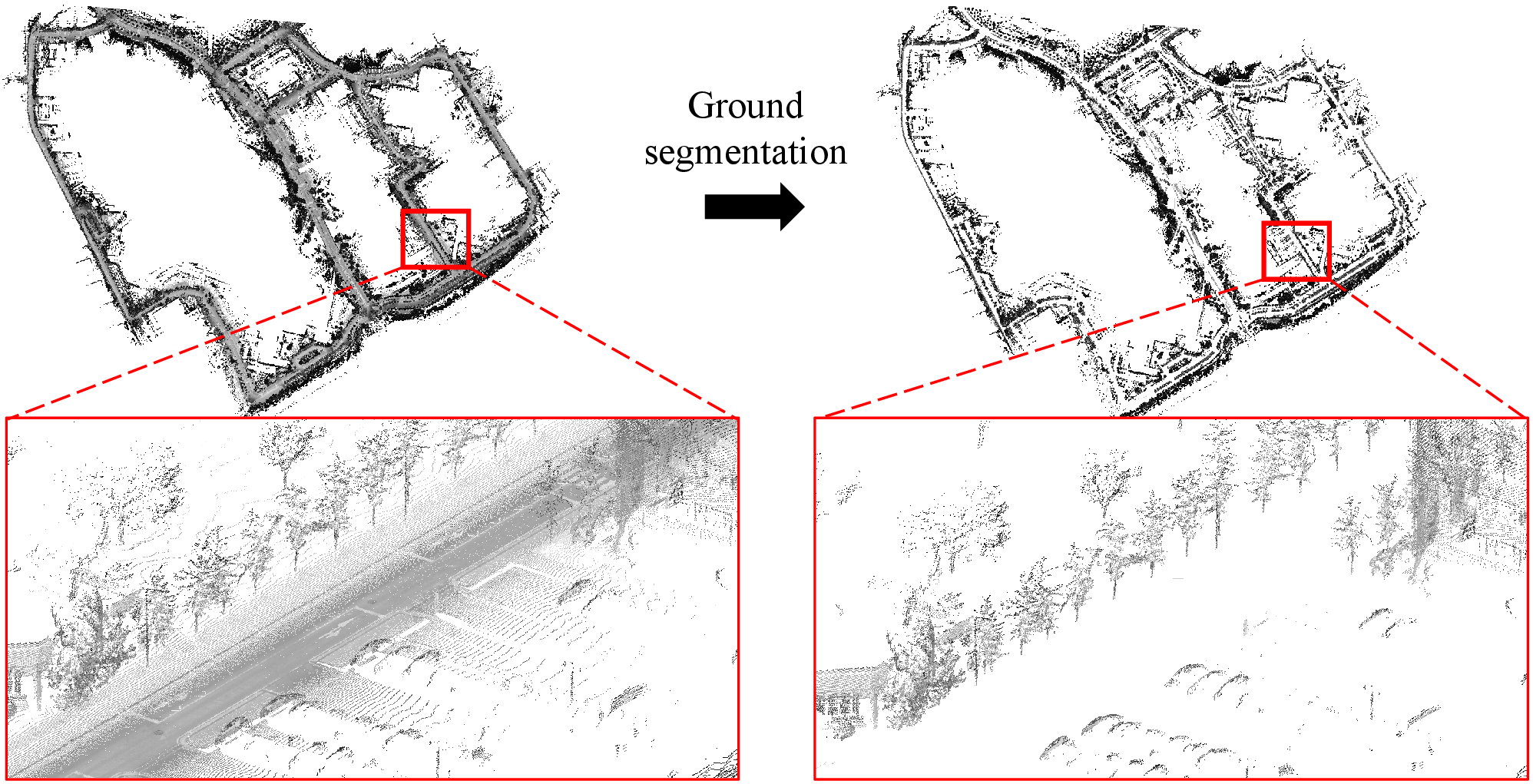}
		\caption{\centering}
	\end{subfigure}
	\vspace{-0.1cm}
	\caption{(a)-(b) Examples of ground segmentation as a preprocessing step from egocentric and mapcentric perspectives, respectively. (a) A result of above-ground segmentation: ground segmentation is applied to reject ground points~(gray points) before the main algorithm and then followed by above-ground segmentation~\cite{oh2022travel}. Points with the same color indicate that these points are segmented into the same object. (b)~Before and after the application of map-level ground segmentation. By leaving non-ground points, ground segmentation allows a robot to perform better localization~(best viewed in color).}
	\label{fig:gseg_application}
\end{figure}

Second, ground segmentation makes non-ground points more distinguishable by rejecting the geometrically featureless points in advance and thus effectively separating non-ground points in 3D space.
For these two reasons, ground segmentation is exploited as a preprocessing step in many applications from both egocentric and map-centric perspectives, as shown in \figref{fig:gseg_application}.

\subsection{Approaches to Ground Segmentation}

Ground segmentation approaches are mainly divided into three types: a) elevation-based~\cite{thrun2006stanley, asvadi2015itsc}, b)~geometric model-based~\cite{fischler1981ransac, douillard2011segmentation, narksri2018slope, moosmann2009iv} and c)~regression-based approaches~\cite{paigwar2020gndnet}.
The common concept of ground segmentation approaches is region-wise estimation, which assumes that points in a small local region are sufficiently flat.

Based on this assumption, region-wise approaches offer two advantages.
First, compared with singular plane model-based methods~\cite{fischler1981ransac}, these region-wise approaches~\cite{asvadi2015itsc,thrun2006stanley, lim2021patchwork, lee2022patchworkpp, oh2022travel, himmelsbach2010fast, steinhauser2008motion, narksri2018slope, cheng2020simple} allow us to overcome the non-planarity of partially steep slopes, bumpy roads, and urban structures that make some regions uneven, \eg~curbs or flower beds.
Second, the total computational cost is reduced by separating a large number of points into multiple small subsets and then estimating the ground points from a small number of points multiple times~\cite{lee2022patchworkpp}.
For these reasons, grid representation-based methods have been widely utilized in contrast to singular plane model-based methods.

In particular, when partitioning a 3D point cloud into multiple grids~(or bins), a polar grid representation~\cite{kim2018scancontext, lim2021erasor, lim2021patchwork}, which divides a point cloud along the radial and azimuthal directions, has been exploited.
This is because the polar grid considers the characteristics of 3D LiDAR sensors that measure points in cylindrical coordinates; thus, the polar grid representation naturally compensates for the sparse characteristics of a 3D point cloud~\cite{himmelsbach2010fast, steinhauser2008motion, narksri2018slope, cheng2020simple}.

Conventional region-wise approaches are mainly classified into a)~elevation map--based and b)~model fitting--based methods.

\subsubsection{Elevation-Based Region-Wise Ground Segmentation}

Elevation map-based methods classify points in a grid as ground points by following two conditions~\cite{thrun2006stanley, asvadi2015itsc}.
By denoting the sensor height from the ground as $h$, the first condition arises from the fact that the $z$~(height direction) values of the actual ground points with respect to the sensor frame are likely to be $-h$.
Second, it is assumed that ground planes are continuous, such that robots can easily traverse these regions compared with the regions with large $z$ value differences.
Thus, the points in a grid are considered as ground points if their average height is close to the negative value of the sensor height and if the value is similar to the average elevation of their neighboring grids.

However, both conditions rely heavily on the premise that the ground is mostly flat, which corresponds to the leftmost case shown in~\figref{fig:categorization}.
Hence, the initial stage of ground segmentation is not strictly distinguished from traversability estimation~\cite{gennery1999traversability, thrun2006stanley , asvadi2015itsc}.
Furthermore, these map-based methods sometimes erroneously estimate steep slope regions as non-ground owing to their relatively high elevation values compared with those from completely flat and planar regions.

\subsubsection{Geometric Model Fitting-Based Region-Wise Ground Segmentation}

To precisely discern between ground and non-ground points, Himmelsbach~\etalcite{himmelsbach2010fast} and Steinhauser~\etalcite{steinhauser2008motion} proposed 2D line fitting-based approaches.
By reducing the size of each grid to be much smaller than those of the elevation-based approaches, a few points originating from a single channel of the LiDAR sensors were assigned to each grid and then line fitting was applied.
Finally, based on the point-to-line distances, gradients, and magnitude of the intercept, these points are decided whether they are ground or not.

However, 2D line fitting-based approaches are susceptible to noise.
Thus, recent ground segmentation approaches have introduced region-wise plane fitting to enhance the robustness against noise~\cite{zermas2017fast, narksri2018slope, cheng2020simple, lim2021patchwork, lee2022patchworkpp}.
For example, Zermas~\etalcite{zermas2017fast} estimated three ground planes of the front, middle, and rear regions along the forward direction of a robot.
However, this approach assumes that changes in the slope along the forward movement of a robot are likely to happen, which may often be violated by rough terrain or complex intersections.
For robustness against slopes, Narksri~\etalcite{narksri2018slope} proposed slope-robust cascaded ground segmentation by reusing the estimates from grids that are closer to the origin when fitting ground planes for the farther grids.
Narksri~\etalcite{narksri2018slope} and Cheng~\etalcite{cheng2020simple} adaptively adjusted grid size based on the density of cloud points or incidence angles.

As an improved version of these plane fitting-based approaches, Patchwork~\cite{lim2021patchwork} was proposed to efficiently check the ground likelihood in terms of uprightness, elevation, and flatness.
Lee~\etalcite{lee2022patchworkpp} proposed Patchwork++, an adaptive approach that automatically updates parameters to adjust to the environments.

\subsubsection{Regression-Based Ground Segmentation} \label{subsec:regression}

Furthermore, several studies have focused on regression-based approaches.
For instance, Douillard \etalcite{douillard2011segmentation}, Chen \etalcite{chen2014gaussian}, and Mehrabi~\etalcite{mehrabi2021gaussian} employed a Gaussian process~(GP) regression and repeated this GP process until convergence.
These methods can discern ground points more precisely; however, GP-based approaches require more computational time than that for fitting-based approaches.

On the other hand, using deep learning~\cite{simonyan2014very, long2015fully,milioto2019rangenet++, choy20194d, paigwar2020gndnet}, point-wise prediction can be easily achieved by supervised learning-based approaches.
For instance, Milioto~\etalcite{milioto2019rangenet++} proposed RangeNet++, which predicts point-wise classes, such as roads, sidewalks, vegetation, and vehicles.
Paigwar~\etalcite{paigwar2020gndnet} presented a network specialized for ground segmentation, called GndNet, which estimates the elevation information of ground planes for each grid.
However, these methods often require graphic processing units~, which are computationally expensive resources.
Moreover, deep learning-based methods are highly likely to be overfitted to the training dataset.
Consequently, their performance may significantly degrade when applied to different environments or sensor configurations from a training dataset~\cite{wong2020identifying}.

\section{TRAVERSABILITY ESTIMATION}

\subsection{Objective of Traversability Estimation}

Compared with ground segmentation, the goal of traversability estimation is to understand the surroundings at a high level of abstraction;
thus, it is not easy to explicitly represent the objective using a simple formula~\cite{beycimen2023survey}.
Specifically, traversability is more complex than ground segmentation, whose role is to divide a 3D point cloud into only two categories,
because traversability is affected by not only~the geometric context of the surrounding environment, \eg~slopes or height discontinuities~\cite{oh2022travel}, but also the kinematic model of a robot~\cite{zou2014comparison}.
This platform-dependent characteristic is further discussed in Section~\rom{4}.

When a robot navigates off-road environments, where complex and irregular objects exist in its surroundings, it becomes more difficult to determine which regions are traversable or not.
For this reason, traversabilty estimation is crucial for robot navigation in off-road scenarios, including agriculture~\cite{reina2017terrain, reina2018allterrain}, planetary exploration~\cite{sancho2010survey, matthies2007computer, bai2019deep, brooks2012self, chhaniyara2012terrain, swan2021ai4mars}, mining~\cite{yu2010digital}, rescue~\cite{hwangbo2019learning, miki2022learning}, etc.

\subsection{Approaches to Traversability Estimation}

According to Beycimen~\etalcite{beycimen2023survey},
traversability estimation can be further classified into a)~terrain classification and b)~traversability analysis.
In terrain classification, researchers aim to predict terrain types to utilize their semantic information~\cite{angelova2007fast,bajracharya2009autonomous,moghadam2009online,best2013terrain,zou2014comparison,hang2017optimum, filitchkin2012feature, lee2011terrain, thomas2015terrain, qi2017cvpr, qi2017pointnet-plusplus}.
For instance, once a robot can discriminate between paved roads, asphalt, and unpaved roads in the surroundings by exploiting classification,
the robot selects less risky paths so that it can safely navigate through paved roads.
Moreover, if the robot recognizes negative obstacles~\cite{borges2022survey}, which lie below ground level, such as depressions and pits,
the robot can re-plan a feasible path around the obstacles, thus preventing navigation failure.
Therefore, terrain classification can be considered as an extended concept of ground segmentation into a multi-class framework in that
it involves classifying terrain into various classes, each of which could be an element of the ground.

\begin{figure}[t!]
	\centering
	\captionsetup{font=footnotesize}
	\begin{subfigure}[b]{0.45\textwidth}
		\includegraphics[width=1.0\textwidth]{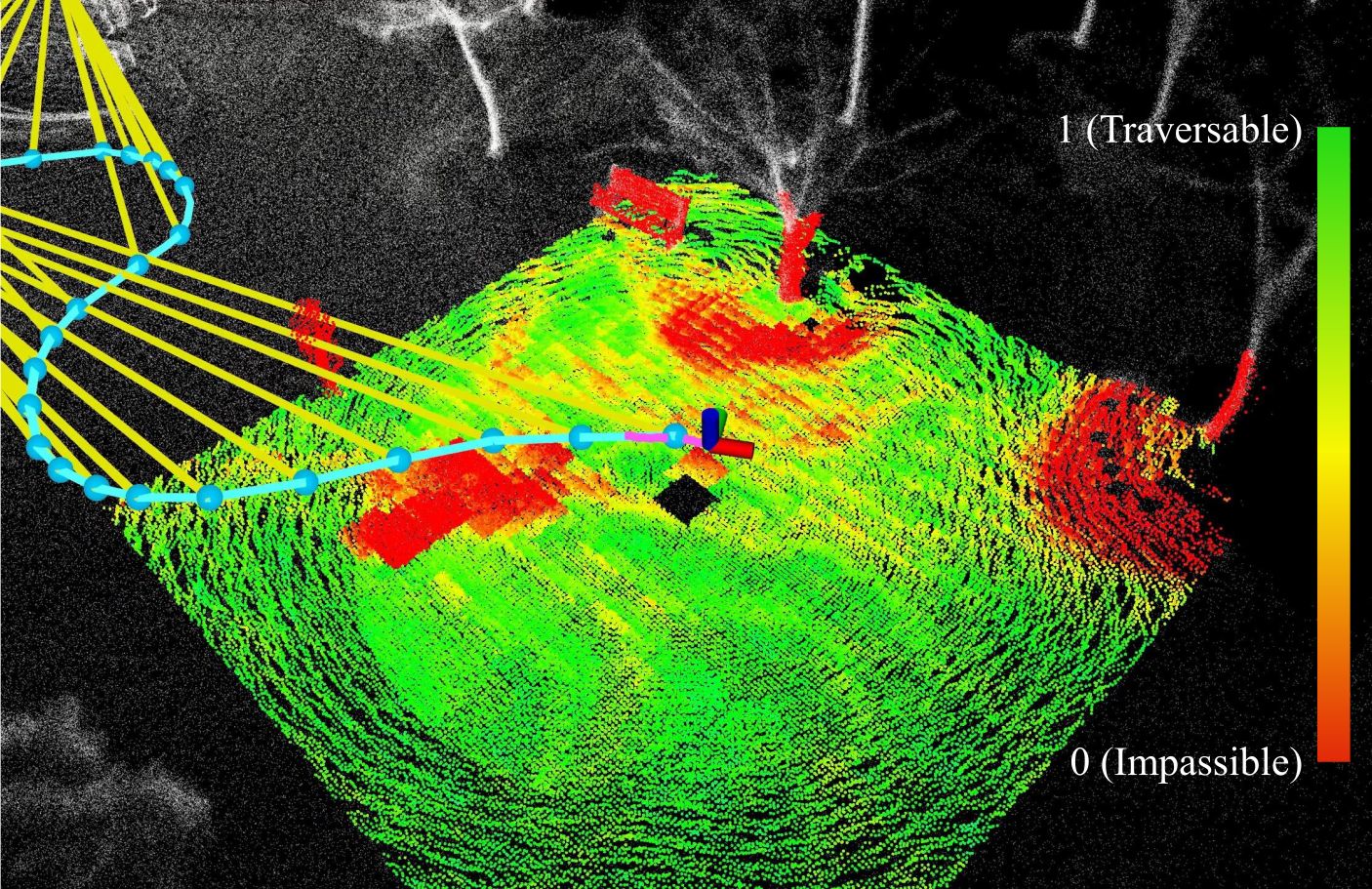}
		\caption{\centering}
	\end{subfigure}
	\begin{subfigure}[b]{0.45\textwidth}
		\includegraphics[width=1.0\textwidth]{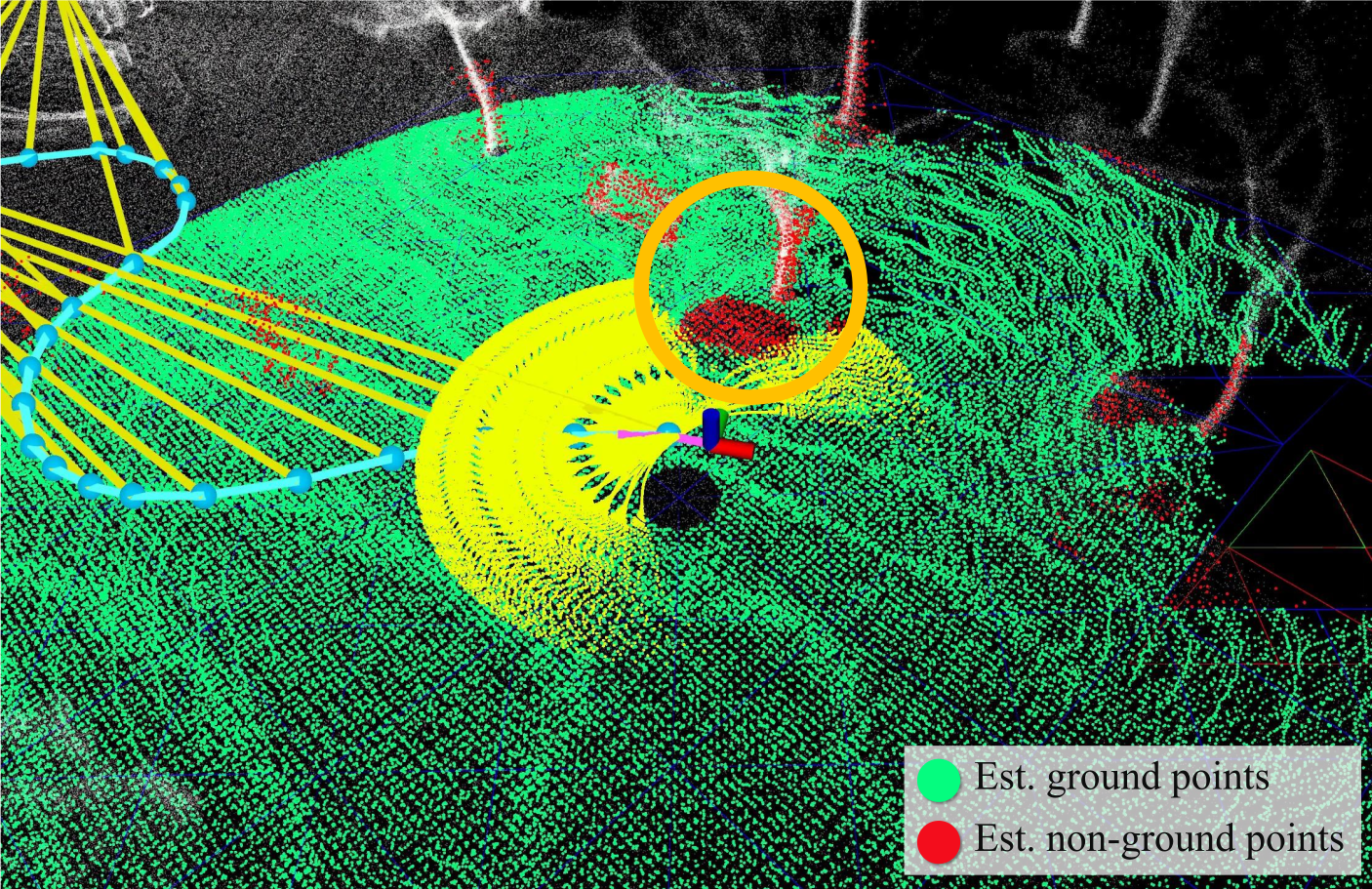}
		\caption{\centering}
	\end{subfigure}
	\caption{(a)~Visual description of the difficulty of traversability, ranging from green to red. The green color represents the regions that are easy to traverse, whereas the red color represents the regions that are impossible to traverse. (b)~Example of path planning~\cite{lee2021real}. The path candidates~(fan-shaped yellow lines around the frame) are not generated near the trunks of trees~(orange circle) by considering the traversability~(best viewed in color).}
	\label{fig:traversability}
	\vsfig
\end{figure}

In contrast to terrain classification, whose primary goal is to recognize the type of an object expressed as a single point or pixel originating from the surroundings,
traversability analysis interprets the data at a more abstract level and then expresses the traversability of the region (or a cell) from zero to one, as shown in~\figref{fig:traversability}.
Here, traversability is defined as the difficulty of a robot to navigate~\cite{papadakis2013terrain}.
Thus, the output of a traversability analysis is generally represented as a 3D point comprising the cost value for each cell.
Because traversability analysis assesses the mobility in a given terrain,
it should account for various aspects, including the local geometry of the region, roughness, friction, and robot kinematics.

\subsubsection{Terrain Classification}
Vision-based sensors are mostly exploited to achieve successful terrain classification~\cite{angelova2007fast,bajracharya2009autonomous,moghadam2009online,best2013terrain,zou2014comparison,hang2017optimum, filitchkin2012feature, lee2011terrain, wellhausen2019, hirose2018gonet, valada2017deep, iwashita2019tu, rothrock2016spoc, maturana2018real, suryamurthy2019terrain, long2015fully, badrinarayanan2017segnet, chen2018a, paszke2016enet}
because they can easily measure the visual appearance of individual pixels on the image plane, such as edges, intensity, color, and texture, compared with point clouds from LiDAR sensors~\cite{shan2018bayesian, lim2020msdpn}.
However, vision sensors are significantly affected by illumination changes and color shifts~\cite{borges2022survey}.
To address these potential limitations, numerous researchers have proposed robust approaches, which can be categorized into three types: a)~non-learning-based, b)~machine learning-based, and c)~deep learning-based approaches.

Terrain classification using traditional non-learning-based methods relies on handcrafted features~\cite{angelova2007fast} or surface reflectivity captured by a stereo camera~\cite{manduchi2005obstacle}.
Kuthirummal~\etalcite{kuthirummal2011graph} proposed a graph-based approach for finding the traversable, obstacle, and unknown cells by exploiting u-disparity occupancy grid and elevation histogram-based outlier rejection.
Dubbelman~\etalcite{dubbelman2007obstacle} utilized a hysteresis threshold to adapt system operation for both day and night conditions, and multi-resolution-based disparity estimation~\cite{vandermark2006} to overcome the ambiguity of the disparity map.
Bajracharya~\etalcite{bajracharya2009autonomous} deployed a near-infrared illumination system and demonstrated its robustness in experiments involving different weather conditions and natural environments.
Thomas~\etalcite{thomas2015terrain} used an airborne 3D point cloud and classified each point into vegetation, buildings, roads, and so forth based on the maximum likelihood estimation.
Reina~\etalcite{reina2020mindtheground} proposed a power-spectral density-based approach to estimate the roughness of terrains, which can be followed by roughness-aware path planning.

The machine learning-based approaches~\cite{angelova2007fast,bajracharya2009autonomous,moghadam2009online,best2013terrain,zou2014comparison,hang2017optimum, filitchkin2012feature, lee2011terrain} use a decision tree or support vector machine.
Angelova~\etalcite{angelova2007fast} and Zou~\etalcite{zou2014comparison} used composite visual information such as texture and color histograms as inputs for the machine learning approaches.
For instance, Angelova~\etalcite{angelova2007fast} proposed a hierarchical classifier that uses various feature representations, including the average color, color histograms, and texture-based features, by employing a decision tree classifier at each level.
Filitchkin and Byl~\cite{filitchkin2012feature}, and Lee and Kwak~\cite{lee2011terrain} exploited descriptors of speeded-up robust features~\cite{bay2006surf}, as input vectors for training.
Lee and Kwak~\cite{lee2011terrain} employed a shallow multilayer perceptron network.
Hang~\etalcite{hang2017optimum} proposed a bag of words-based fusion method that is robust against diverse noise and illumination changes.

Similar to deep learning-based approaches in ground segmentation fields, as explained in Section~\rom{2}.\textit{B}.3, these machine learning-based approaches are substituted with convolutional neural network~(CNN)-based approaches~\cite{maturana2018real, redmon2013darknet, romera2017erfnet, ronneberger2015unet, badrinarayanan2017segnet, chen2018a, chen2018bdeeplab, bai2019deep,qi2017cvpr, qi2017pointnet-plusplus}.
The nonlinearity of CNNs allows the extraction of better feature descriptors, thereby increasing the distinction of features between different classes.
In addition, as more optimized architectures for handling sparse point clouds are employed~\cite{choy20194d, cortinhal2020salsanext},
many terrain classification~(or segmentation) approaches can easily achieve significant performance improvement by presenting detailed predictions of 3D cloud points.
Furthermore, most deep learning-based approaches are based on supervised learning; thus, they are easily trained using labeled data.
In contrast, machine learning-based approaches require handcrafted feature extraction, but the quality is highly likely to be affected by parameter tuning and domain expertise.

\subsubsection{Traversability Analysis}

Similar to terrain classification, traversability analysis fields have also employed learning-based approaches~\cite{suger2015traversability}; however, learning-based approaches suffer from generalization issues.
In other word, learning-based approaches occasionally show catastrophic failure when there are large discrepancies between the training and test environments.
Furthermore, traversability analysis typically uses sequential sensor data as inputs to map traversable regions by leveraging temporal information, which triggers the network to be potentially overfitted to the training dataset.
For these reasons, compared with terrain classification fields, traversability analysis employs various conventional approaches~\cite{krusi2017driving, ruetz2019ovpc, langer1994behavior, gennery1999traversability, hamner2008efficient},
considering 2.5D elevation information~\cite{fankhauser2018probabilistic} or region-wise slopes~\cite{zhang2022vision}.

For instance, Langer~\etalcite{langer1994behavior} and Gennery~\etalcite{gennery1999traversability} exploited the variances of point distributions in each cell to check the roughness.
Wermelinger~\etalcite{wermelinger2016navigation} proposed a traversability map, whose traversability was estimated by the roughness, height, and local slope, and then employed this traversability information to perform path planning in challenging terrains.
Kim~\etalcite{kim2020vision} leveraged the concept of steppability, which indicates whether it is geometrically possible for a legged robot to be stepped on with its foot, to execute footstep-level planning.
Xue~\etalcite{xue2023jfr} used local convexity to estimate traversable regions, which are obtained from the dense elevation and surface normal maps.

To manage the sparse characteristics of a 3D point cloud captured by a LiDAR sensor, probability-based terrain inference approaches have also been proposed to densely fill the holes in the cells.
Shan~\etalcite{shan2018bayesian} applied a Bayesian generalized kernel-based inference approach that infers the traversability cost of an empty cell using the costs of neighboring nonempty cells.


\section{Comparison Between Ground Segmentation and Traversability Estimation}\label{sec:comparison}

In this section, we claim that the targets of ground segmentation and traversable region detection can be differentiated even though the same robot is employed in the same environment.
For example, from the perspective of autonomous vehicles that navigate in on-road environments, traversable regions are mostly roads,
while the ground includes all areas where an object can come in contact with, as shown in \figref{fig:taxonomy}.
This difference raises a question of \textit{what factors influence the categories of ground and traversable regions}.

To answer this question, we set four following criteria: a)~maneuverability of platforms, b)~position in the surroundings, c)~subset relation of negative obstacles, and d)~subset relation of deformable objects, as summarized in \tabref{table:gs_te}.

\subsection{From the Perspective of Platforms}\label{subsec:platforms}

\begin{figure}[t!]
	\centering
	\captionsetup{font=footnotesize}
	\begin{subfigure}[b]{0.15\textwidth}
		\includegraphics[width=1.0\textwidth]{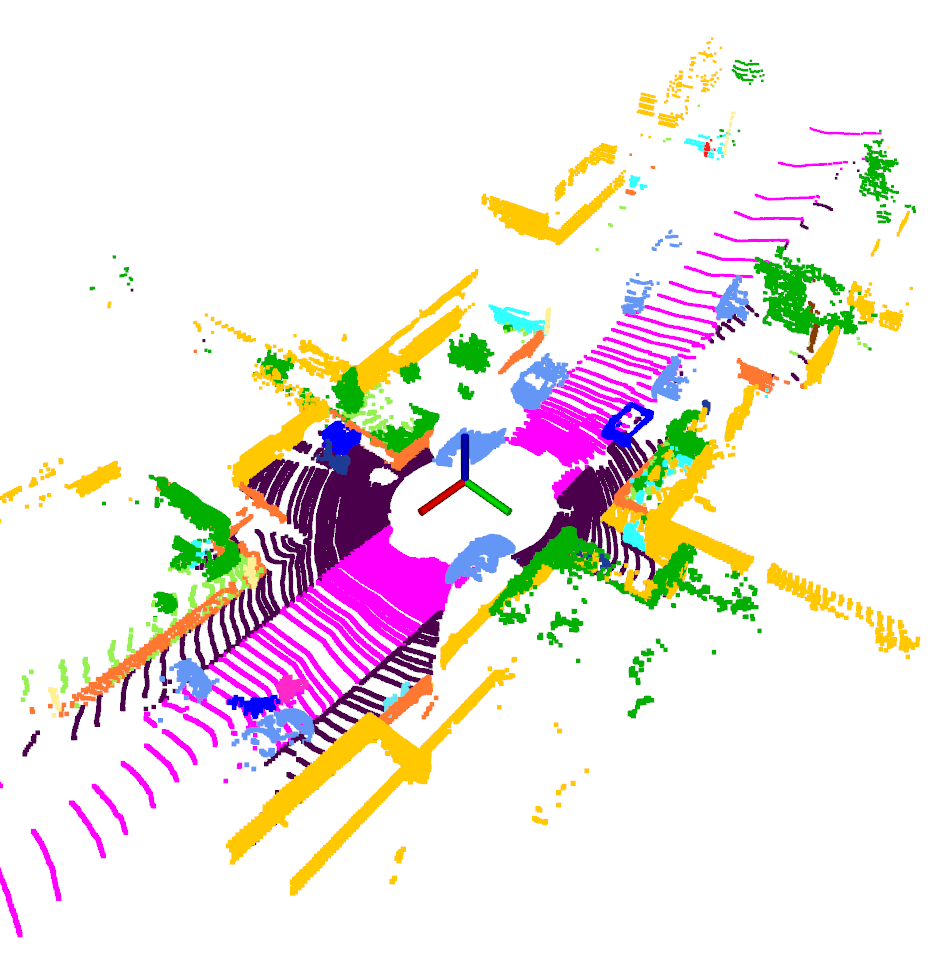}
		\includegraphics[width=1.0\textwidth]{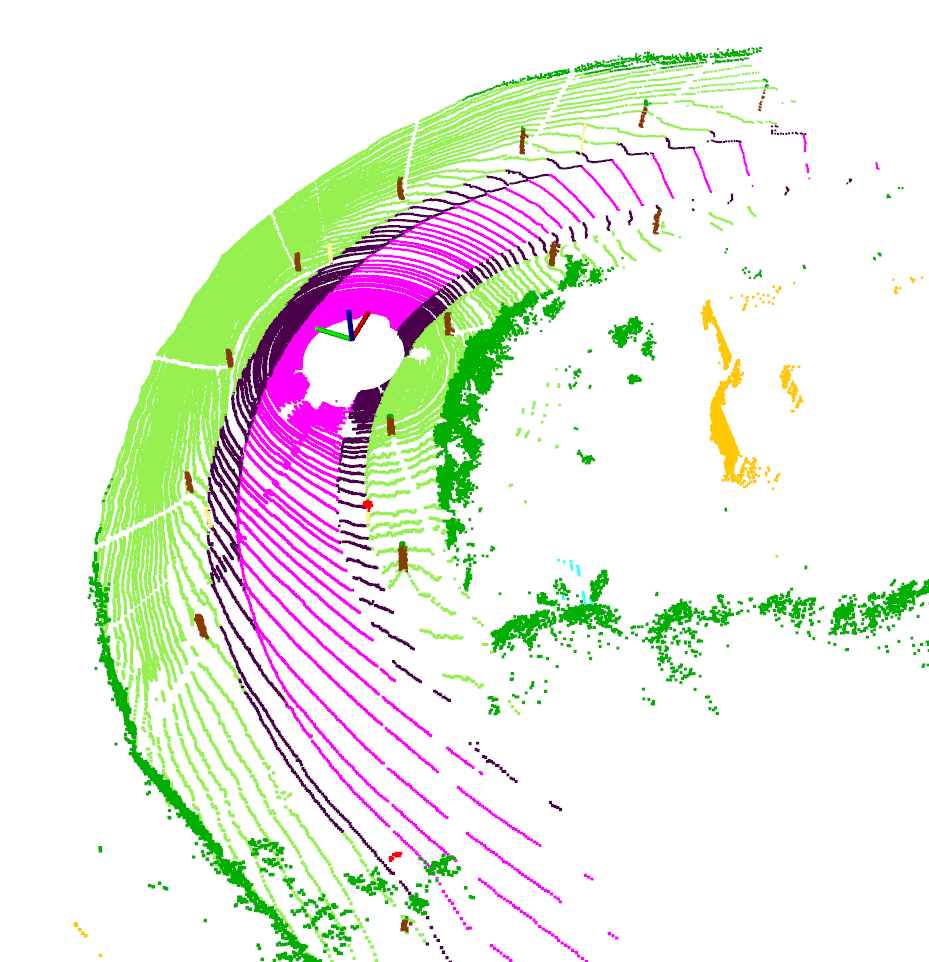}
		\caption{\centering}
	\end{subfigure}
	\begin{subfigure}[b]{0.15\textwidth}
		\includegraphics[width=1.0\textwidth]{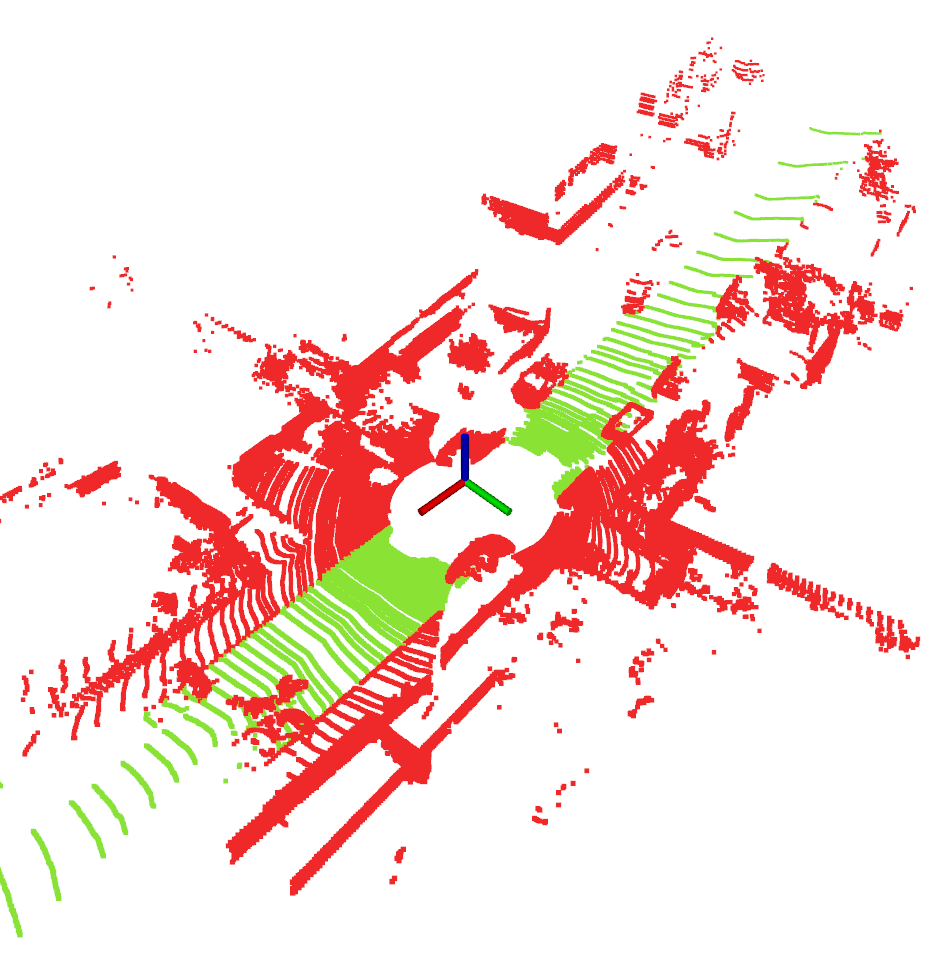}
		\includegraphics[width=1.0\textwidth]{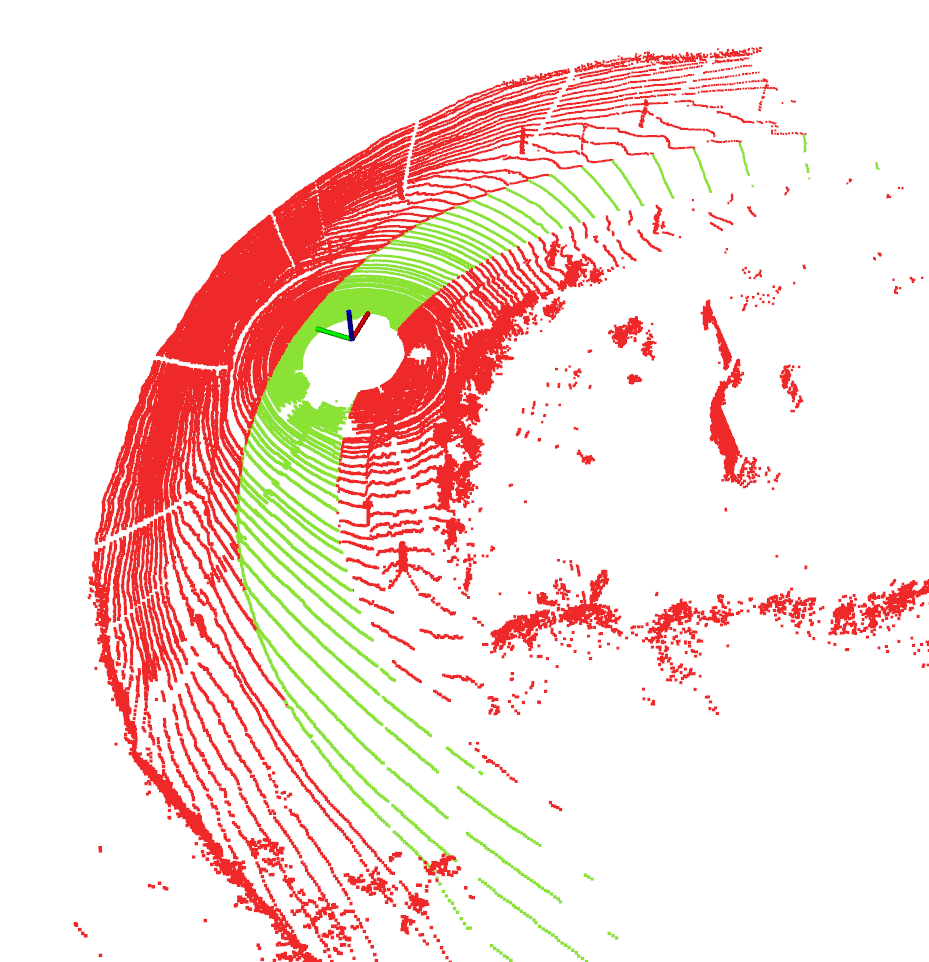}
		\caption{\centering}
	\end{subfigure}
	\begin{subfigure}[b]{0.15\textwidth}
		\includegraphics[width=1.0\textwidth]{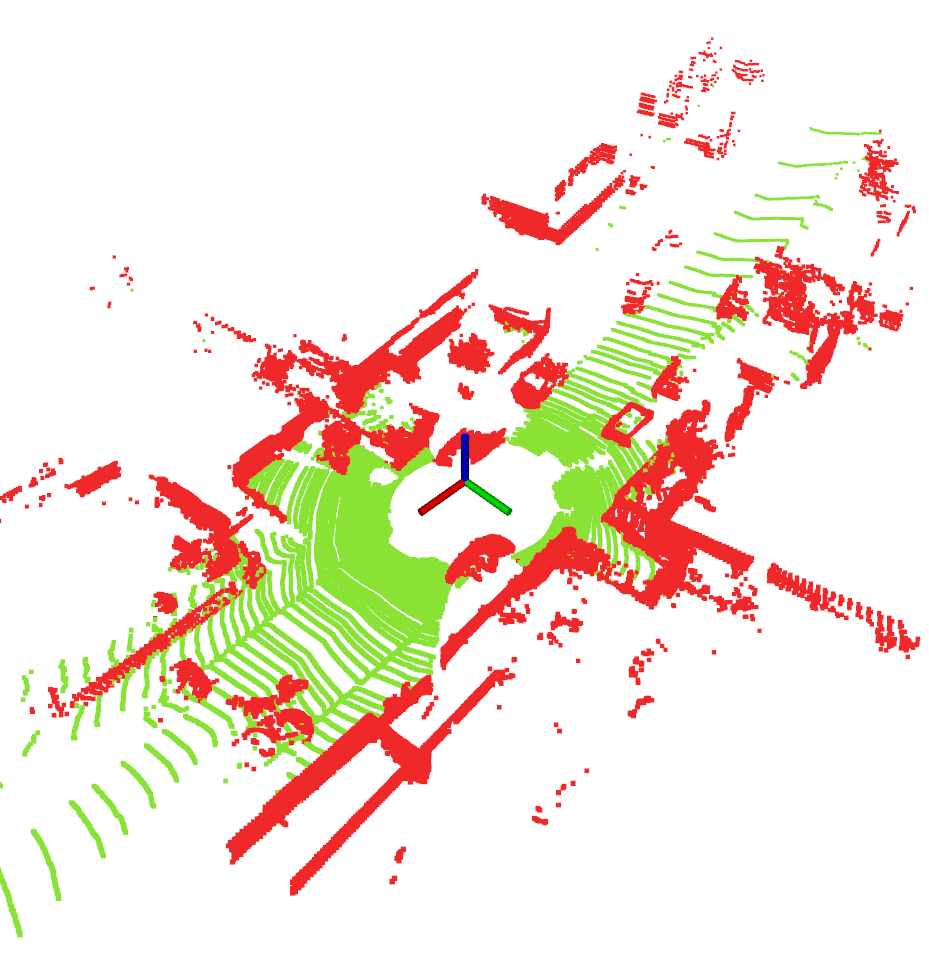}
		\includegraphics[width=1.0\textwidth]{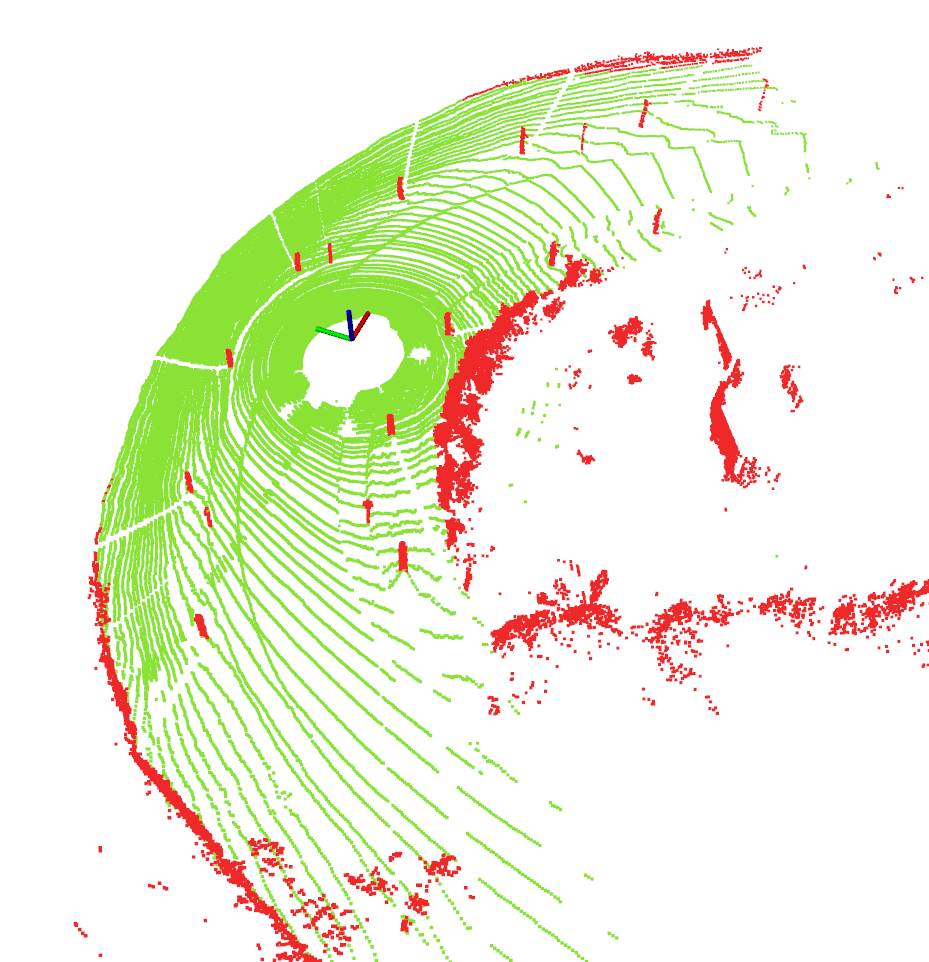}
		\caption{\centering}
	\end{subfigure}
	\caption{Visual description of how traversable and ground regions can be differentiated depending on the purposes. (a)~Visualized SemanticKITTI labels~\cite{behley2019iccv}, where light green, dark brown, and magenta colors represent \texttt{terrain}, \texttt{sidewalk}, and \texttt{road} classes, respectively (these colors follow the SemanticKITTI-API). (b)~Traversable regions in terms of traversable area detection and (c)~ground in terms of ground segmentation from the perspective of a wheeled robot. In (b) and (c), the light green color represents our targets of interest, \ie~traversable and ground regions, respectively, and the red colors represent their complementary sets~(best viewed in color).}
	\label{fig:taxonomy}
	\vsfig
\end{figure}

First, we stress that the ground is a platform-agnostic concept, whereas traversable regions are highly affected by the moving mechanisms of robots.
Even though the type of the robot used for acquiring the point cloud in \figref{fig:taxonomy}~(originally, an autonomous vehicle was used) is changed to a large amphibious unmanned ground vehicle~(UGV) or quadrupedal robot, the ground regions are not changed~(\figref{fig:taxonomy}(c)).
This is because the definition of non-ground points, which are objects of interest, does not change because non-ground objects in the surrounding environment are irrelevant and invariant to the robot platform, as described in 

In contrast, the traversable regions can change depending on the robot platforms.
For example, assuming that we employ a large amphibious UGV or quadrupedal robot, the sidewalk can also be considered as a traversable region because the robot with high maneuverability crosses the curbs which make the regions between the sidewalks and roads discontinuous.
Hence, we conclude that the traversable regions are platform-dependent.

\subsection{From the Perspective of Position}

\begin{figure}[t!]
	\centering
	\captionsetup{font=footnotesize}
	\begin{subfigure}[b]{0.23\textwidth}
		\includegraphics[width=1.0\textwidth]{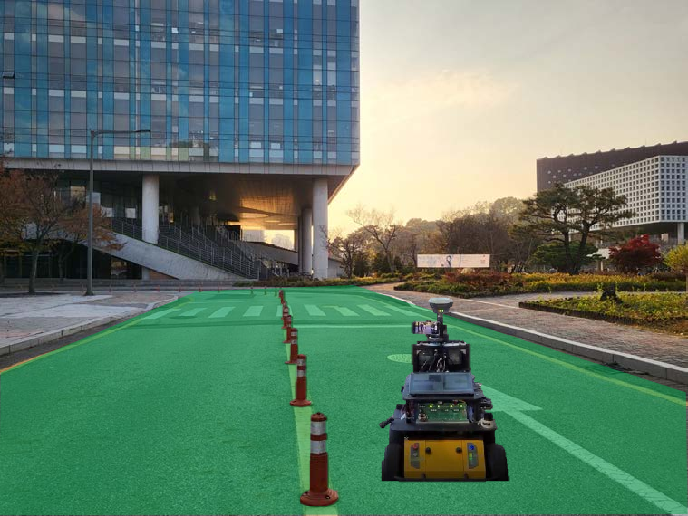}
		\includegraphics[width=1.0\textwidth]{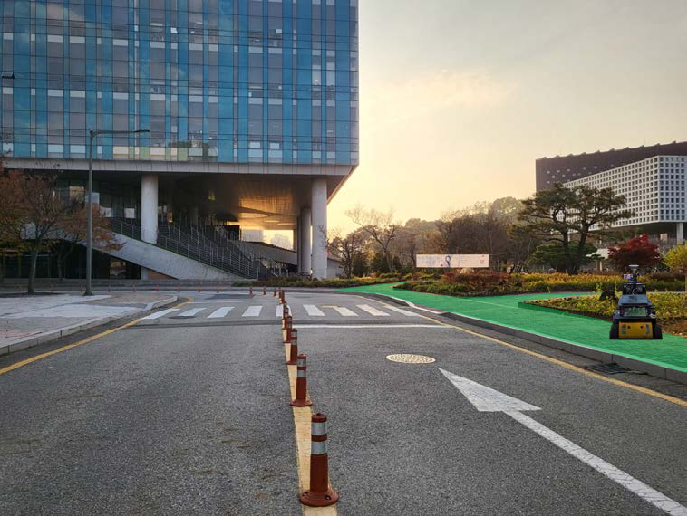}
		\includegraphics[width=1.0\textwidth]{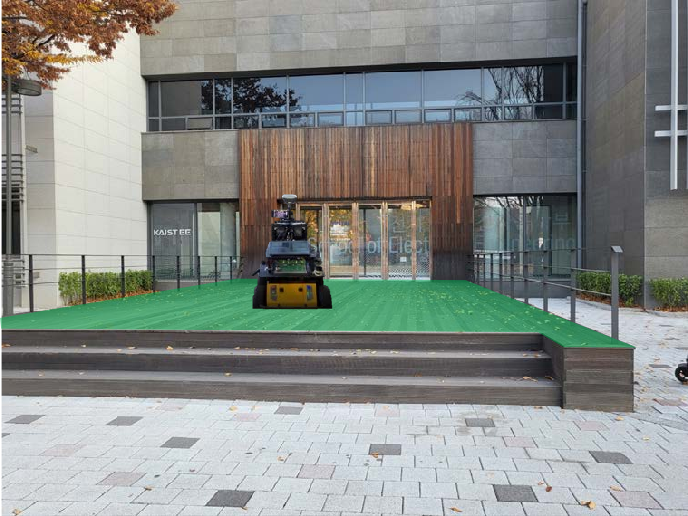}
		\caption{\centering}
	\end{subfigure}
	\begin{subfigure}[b]{0.23\textwidth}
		\includegraphics[width=1.0\textwidth]{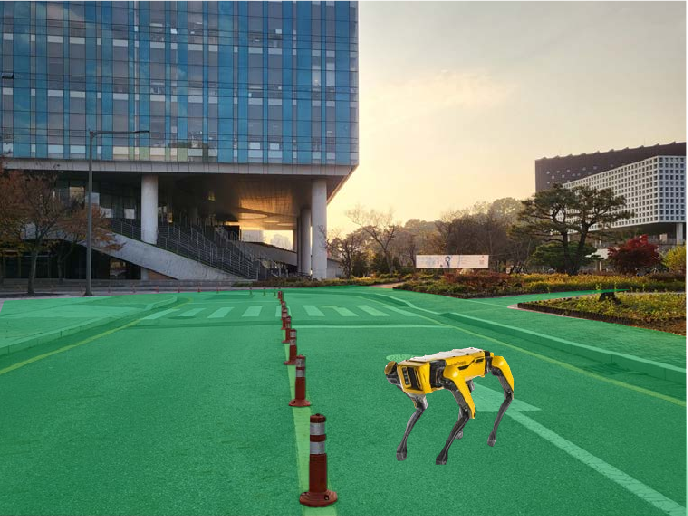}
		\includegraphics[width=1.0\textwidth]{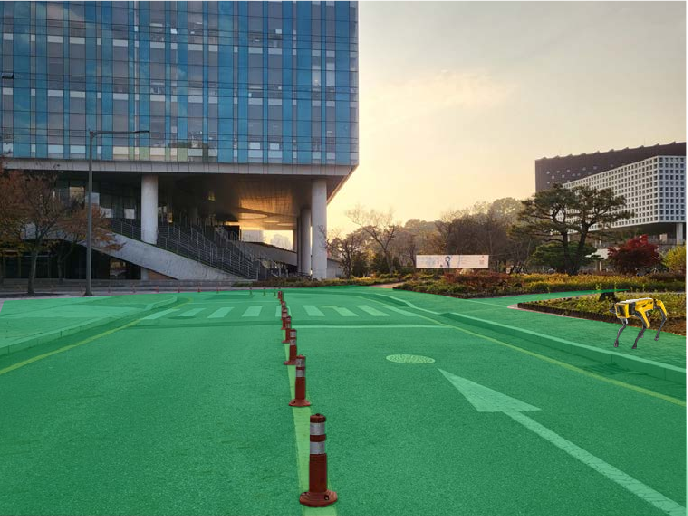}
		\includegraphics[width=1.0\textwidth]{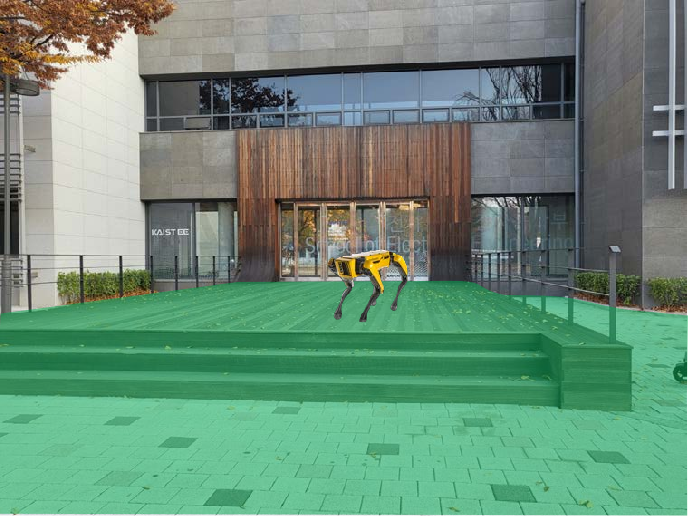}
		\caption{\centering}
	\end{subfigure}
	\vspace{-0.1cm}
	\caption{Visual description of traversable regions depending on the maneuverability and positions of the (a)~wheeled and (b)~quadruped robots in urban environments, respectively. The green colors represent the traversable regions for the robot located in that position~(best viewed in color).}
	\label{fig:position}
	\vsfig
\end{figure}

As described in Section~\rom{4}.\textit{A}, the traversable regions are also position-dependent owing to the limited mobility of the robot.
As shown in~\figref{fig:position}(a), in urban environments, the traversable regions can vary depending on the position of the wheeled robot because it is difficult for the robot to navigate through curbs and stairs,
whereas the walking mechanism enables legged robots to easily overcome these small obstacles; thus, wider areas can be considered traversable.

From the viewpoint of ground segmentation, the ground remains invariant to the positions because regardless of the location of the robot, the definition of non-ground objects does not change.
In this respect, the ground can be all areas where a human or car can come into contact.
Therefore, the ground mostly includes traversable regions for the robots.

\input{contents/comparison.tex}

\subsection{From the Perspective of Negative Obstacles} \label{subsec:neg_obstacles}

Next, we investigate how negative obstacles~\cite{borges2022survey}, which refer to obstacles that lie below ground level, affect ground segmentation and traversability estimation.
Regarding ground segmentation, the points from these negative obstacles can only be considered as ground points because only the points from positive obstacles, such as trees, walls, humans, fences, and vehicles, are considered as non-ground points.
However, regarding traversability, the regions occupied by negative obstacles should be avoided when passing through because negative obstacles may hinder robots from successful navigation owing to the wheels or legs getting stuck.

In contrast, when the robot has long legs or the radius of its wheels is sufficiently large, the regions where negative obstacles exist are also considered as traversable regions.
Consequently, as presented in~\tabref{table:gs_te}, the subset relation between negative obstacles and traversable regions can differ depending on the maneuverability of robots.

\subsection{From the Perspective of Deformable Objects}\label{subsec:deformable}

Finally, deformable objects require clarification on whether they are included in the ground or traversable regions.
Deformable objects, such as lawns, bushes, and reeds, can be folded or trampled.
For this reason, the categorization of deformable objects is ambiguous because whether deformable objects are ground or traversable regions is also determined by the size of the robots.
In other word, depending on the mobility of the robot, deformable objects may or may not be included in the ground.
Thus, in some cases where a wheeled mobile robot is employed, large vegetation can be considered non-ground objects because it makes the ground surface occupied and discontinuous.

\begin{figure}[t!]
	\centering
	\captionsetup{font=footnotesize}
	\begin{subfigure}[b]{0.23\textwidth}
		\includegraphics[width=1.0\textwidth]{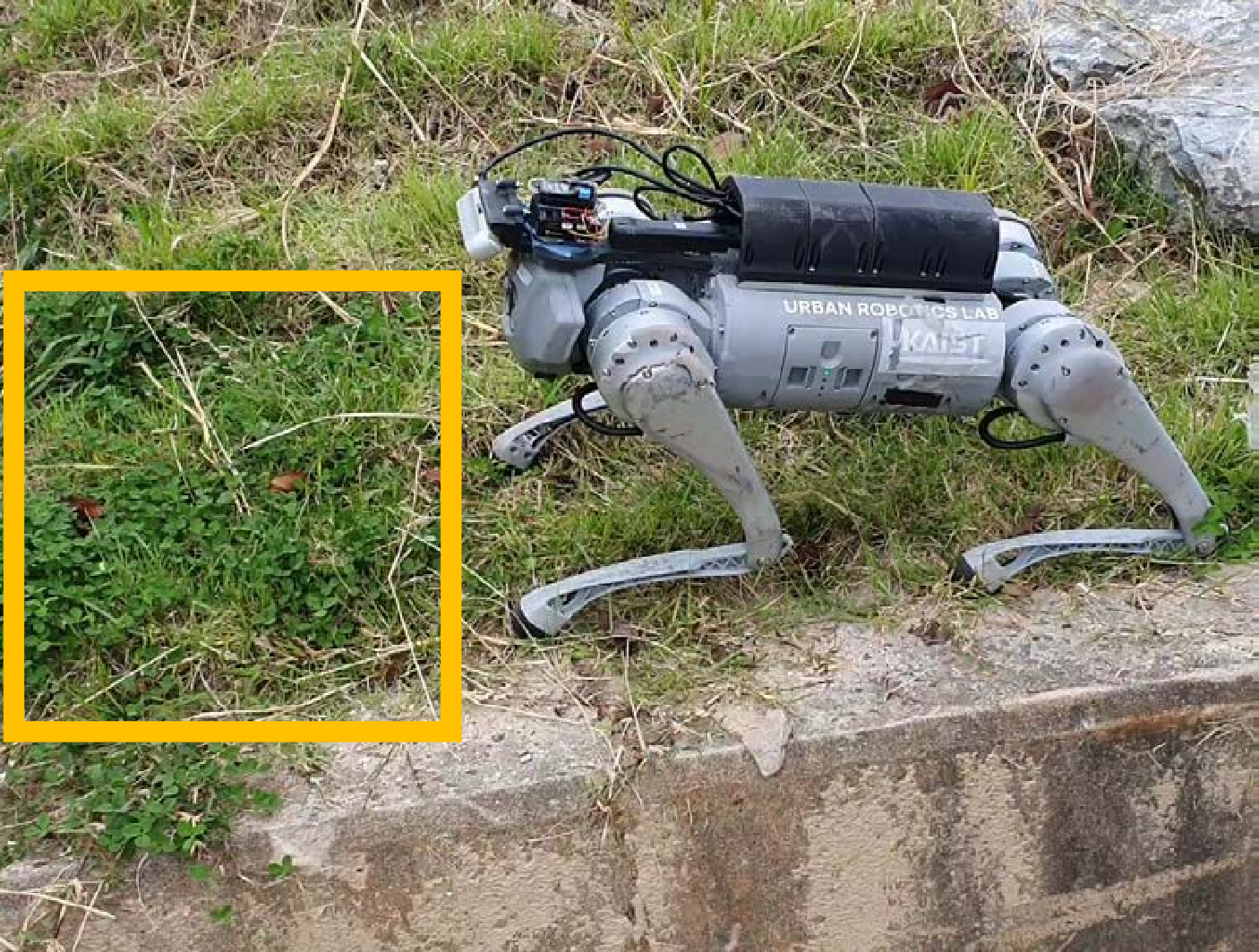}
		\caption{\centering}
	\end{subfigure}
	\begin{subfigure}[b]{0.23\textwidth}
		\includegraphics[width=1.0\textwidth]{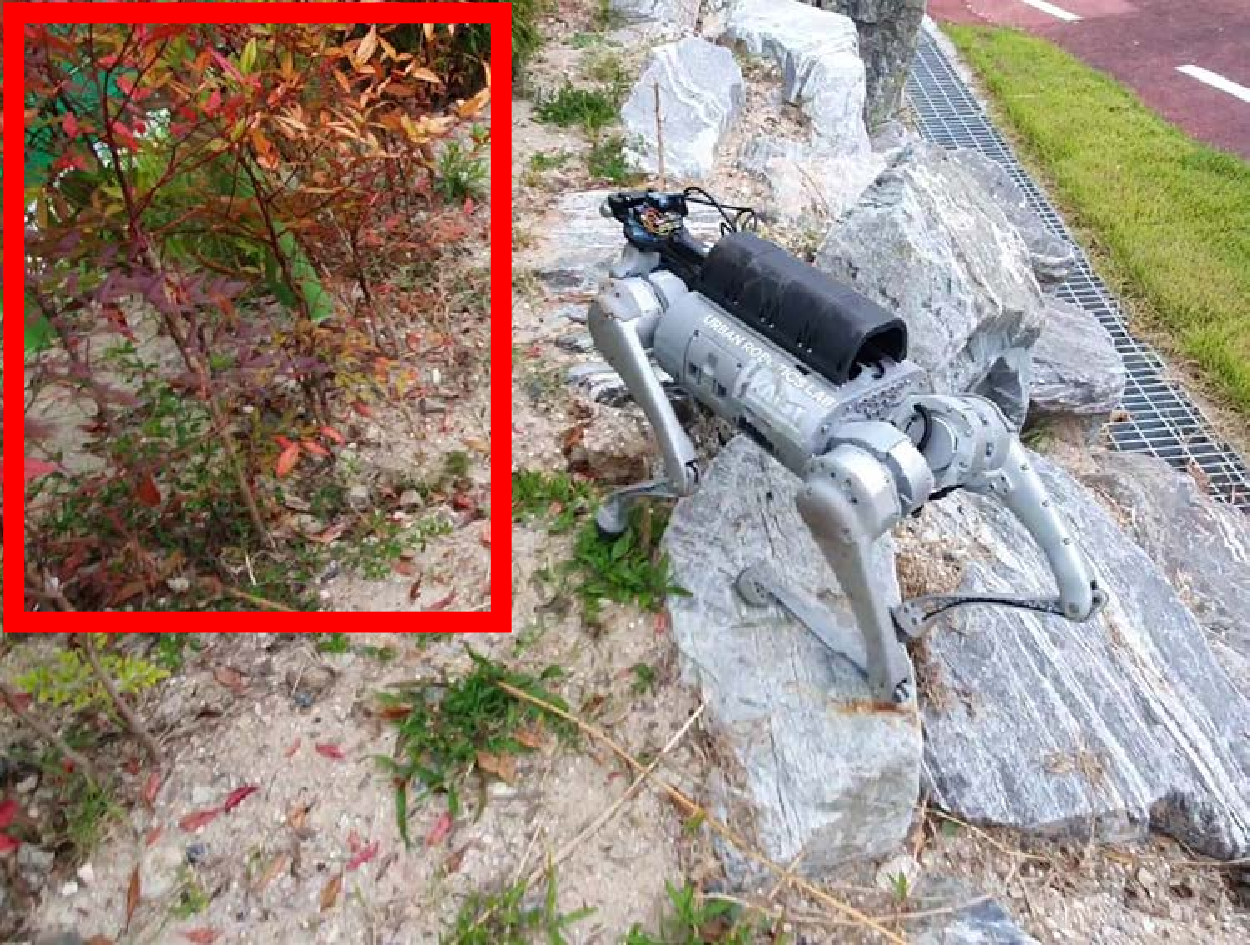}
		\caption{\centering}
	\end{subfigure}
	\vspace{-0.1cm}
	\caption{Examples of (a) small deformable objects~(lawns in the orange box) and (b) large deformable objects~(reed-like vegetation in the red box) from the perspective of a small quadruped robot.}
	\label{fig:deformable}
	\vsfig
\end{figure}

Unlike a small wheeled mobile robot that considers vegetation as a definite obstacle,
a quadruped robot can traverse lawns and reed-like vegetation, albeit with some stumbling, as shown in~\figref{fig:deformable}.
In this respect, similar to the negative obstacles described in Section~\rom{4}.\textit{C}, deformable objects are also highly affected by the maneuverability of a robot.
Therefore, the subset relation can vary depending on the relative sizes of the deformable objects and the robot.

\section{Conclusion}

In this paper, we explained the concept of the ground segmentation and traversability estimation and then clarified the distinction between ground and traversable regions.
We conclude that ground segmentation divides data into ground and non-ground elements at the perception level as a preprocessing stage, whereas traversability estimation identifies and comprehends areas where robots can move safely at the cognition level.
By exploring these differences, we have found that the ground is less affected by external factors, including robot platforms and positions in the surroundings.
In contrast, traversable regions are more likely to vary depending on the maneuverability of the robot platforms.
Therefore, we hope that other researchers will differentiate and use these terminologies based on whether they are intended for perception-level preprocessing or path planning at the cognitive-level.

\bibliographystyle{IEEEtran}
\bibliography{main.bib}

\end{document}

%% file: contents/comparison.tex
\begingroup
\begin{table}[t!]
    \captionsetup{font=footnotesize}
	\centering
	\caption{Summary of the differences between ground segmentation and traversability estimation in terms of five aspects. Note that the size of deformable objects is a relative concept determined by the size of the robot. The symbols $\bigcirc$ and $\times$ denote the targets that are most likely to be included and not included, respectively; $\triangle$ represents that the subset relation depends on the maneuverability of a robot.}
	{\scriptsize
	\begin{tabular}{c|cc}
	\toprule \midrule
	Criteria & Ground segmentation & Traversability estimation \\ \midrule
	Platform & Agnostic & Dependent \\ \midrule
	Position & Agnostic & Dependent \\ \midrule
	\begin{tabular}{@{}c@{}}Negative \\ obstacles  \end{tabular} & $\bigcirc$ & $\triangle$ \\ \midrule
	\begin{tabular}{@{}c@{}}Small \\ deformable \\ objects \end{tabular}  & $\bigcirc$ & $\triangle$ \\ \midrule
	\begin{tabular}{@{}c@{}}Large \\ deformable \\ objects \end{tabular}  & $\times$ & $\triangle$ \\
	\midrule\bottomrule
	\end{tabular}
	}
	\label{table:gs_te}
 	\vspace{-0.3cm}
\end{table}
\endgroup